\documentclass[a4paper,fleqn]{cas-dc}

\usepackage[numbers]{natbib}

\def\tsc#1{\csdef{#1}{\textsc{\lowercase{#1}}\xspace}}
\tsc{WGM}
\tsc{QE}
\tsc{EP}
\tsc{PMS}
\tsc{BEC}
\tsc{DE}

\begin{document}
\let\WriteBookmarks\relax
\def\floatpagepagefraction{1}
\def\textpagefraction{.001}

\shorttitle{Evaluating and Enhancing Segmentation Model Robustness with Metamorphic Testing}

\shortauthors{S. Mzoughi et~al.}

\title [mode = title]{Evaluating and Enhancing Segmentation Model Robustness with Metamorphic Testing}                      


\author[1]{Seif Mzoughi}
\cormark[1]
\cortext[cor1]{Corresponding author}
\ead{seif.mzoughi@polymtl.ca}
\credit{Conceptualization, methodology, software, validation, formal analysis, investigation, resources, data curation, writing--original draft preparation, visualization}
\author[1]{Mohamed Elshafei}
\credit{Conceptualization, methodology, validation, formal analysis, investigation, writing--original draft preparation, writing--review and editing, visualization, supervision}
\ead{mohamed.elshafei@polymtl.ca}
\author[1]{Foutse Khomh}
\ead{foutse.khomh@polymtl.ca}
\credit{Conceptualization, validation, resources, writing--review and editing, supervision, project administration, funding acquisition}

\affiliation[1]{organization={Department of Computer and Software Engineering, Polytechnique Montreal},
                city={Montreal},
                postcode={H3T 1J4}, 
                state={Quebec},
                country={Canada}}

\begin{abstract}
Image segmentation is a fundamental component of either image processing or computer vision, finding its applications in medical image analysis, augmented reality, and video surveillance, among others. However, the current research is paying too little attention to the robustness of such models, which is actually a factor that easily predisposes the model to adversarial perturbations caused by slight, imperceptible distortions added to the input images. In this work, we leverage Metamorphic Testing (MT) to evaluate and boost Segmentation models robustness. Our key innovation lies in using GA to intelligently evolve and optimize transformation sequences, systematically discovering the most effective combinations of spatial and spectral distortions while maintaining image fidelity.  Our segmentation robustness metamorphic testing approach (SegRMT) generates adversarial examples that maintain the visual coherence of images while adhering to a predefined Peak Signal-to-Noise Ratio (PSNR) threshold, ensuring genuine disruptions. We use the Cityscapes dataset for our experiments, which consists of 5,000 images from diverse stereo video sequences in urban environments across 50 cities. Our findings show that by combining metamorphic testing and a genetic algorithm (GA), our approach can significantly reduce the mean Intersection over Union (mIoU) produced by the DeepLabV3 segmentation model to 6.4\%, while other baseline adversaries decrease mIoU values between 21.7\% and 8.5\%. Other findings indicate that SegRMT and other baseline adversarial training achieve higher performance if training and testing occurred on their separate specific adversarial datasets, with mIoU values up to 73\%. Other findings indicate that SegRMT adversarial training increases the mIoU of a segmentation model to 53.8\% in cross-adversarial testings, while other baseline adversaries only increase mIoU values to between 2\% and 10\% on the SegRMT adversarial testing. This demonstrates that SegRMT effectively enhances the robustness of segmentation models by generating adversarial examples that more accurately simulate real-world distortions. The key advantage of SegRMT lies in its ability to maintain visual coherence while introducing subtle, yet impactful, perturbations that challenge the model in ways other adversarial methods do not. This suggests that SegRMT can be a valuable tool for improving model reliability in applications where consistent performance is crucial despite varying and unpredictable environmental conditions.
\end{abstract}



\maketitle

\section{Introduction}
Despite the extensive literature about deep learning segmentation models—which aim to automatically partition images into distinct meaningful regions— and their applications, little attention has been given to their robustness against adversarial attacks or image distortions so far.
Nowadays,  with the rise of deep learning segmentation models, there is a growing demand for evaluating their robustness, particularly in critical fields like autonomous driving and medical imaging~\cite{rossolini2023real,muller2022towards}. These critical applications require highly reliable models to prevent potentially devastating consequences. For example, accurately segmenting road scenes under diverse conditions in autonomous driving is crucial for ensuring safety.  In medical imaging, for example, it is equally important that the identification and segmentation of the anatomical structures are as accurate as necessary to ensure correct diagnosis and effective treatment planning. Recent studies suggest that adversarial data, whether controlled (e.g., human-induced attacks) or uncontrolled (e.g., distortions), are on the rise and are hindering the robustness of deep learning segmentation models~\cite{kaviani2022adversarial,bouchoucha2023robustness}. Adversarial human-induced attacks involve introducing artificial alterations to input data in order to deceive the models, resulting in inaccurate predictions and compromising their reliability. Conversely, distortions induced by adversarial environments involve uncontrollable factors, such as variations in lighting, occlusions, and sensor noise, which can also impact the model's performance and compromise their reliability. When testing the robustness of models in real-world applications, traditional gradient methods like the Fast Gradient Sign Method (FGSM), Projected Gradient Descent (PGD), and Carlini \& Wagner (C\&W) are commonly used. However, these methods may not fully account for other adversaries, such as distortions caused by adversarial environments~\cite{wang2024uncovering,usynin2023beyond,dong2020greedyfool}. 

Recently, Metamorphic testing (MT) has emerged as a promising approach to combine gradient-based adversarial attacks and real-world distortions to evaluate and enhance models robustness~\cite{bouchoucha2023robustness}. MT simulates real-world disturbances and generates new test cases by systematically modifying existing ones to evaluate model robustness under diverse input conditions, incorporating adversarial attacks and real-world disturbances~\cite{xie2020mettle,santos2020experimental,xie2011testing}. When incorporated with an optimization algorithm, MT provides scalability and optimization advantages to tune the distortions while avoiding corrupting the input data~\cite{bouchoucha2023robustness,ma2022metamorphic,pan2021metamorphic}. Therefore, in this work, we propose a novel optimization-driven approach to segmentation robustness testing. Our key innovation lies in combining metamorphic testing (MT) with genetic algorithms (GA) to intelligently discover and optimize adversarial distortions. Unlike existing approaches that use fixed transformation patterns or random perturbations, our segmentation robustness metamorphic testing approach (SegRMT) leverages GA's evolutionary optimization to systematically explore the space of possible transformations, identifying the most effective combinations of distortions while maintaining image fidelity. Furthermore, we introduce a threshold for Peak Signal-to-Noise Ratio (PSNR), a quality metric that measures the ratio between maximum possible pixel value and distortion noise, to ensure that SegRMT generates adversarial input data within safe limits for distortions and data integrity~\cite{ma2022metamorphic,sheng2024toward}. Otherwise, over-distortion may result in high corruption and diverge the input data away from the norms. We use the cityscapes dataset to evaluate and enhance the robustness of segmentation models across various experiments~\cite{cordts2016cityscapes}. In the first experiment, we demonstrate the effectiveness of SegRMT compared to the traditional gradient-based adversarial attacks. In the second experiment, we fine-tune the segmentation model through adversarial training using training data and adversarial examples generated by SegRMT and traditional gradient-based adversarial attacks. This exposes the model to a wide range of perturbations during training, thus potentially improving its robustness under adversarial conditions and increasing its performance. In the third experiment, we cross-test the robustness of segmentation models between SegRMT and traditional gradient-based adversarial attacks. 

The contribution of this work is twofold and includes: 1) proposing a novel GA-optimized framework for evaluating segmentation model robustness, which systematically discovers effective adversarial transformations through evolutionary optimization, and 2) enhancing the robustness of the model using adversarial training data. We express and validate these contributions by answering the following research questions.:
\begin{itemize}
    \item RQ1) How effective is SegRMT compared to traditional gradient-based adversarial attacks in deceiving a segmentation model?
    \item RQ2) How effectively does SegRMT enhance segmentation model robustness in self-adversarial testing compared to traditional gradient-based attacks?
    \item RQ3) How effectively does SegRMT improve segmentation model robustness in cross-adversarial testing versus traditional gradient-based attacks?

\end{itemize}

Importantly, while previous work has combined metamorphic testing with genetic algorithms for applications in software testing \cite{10.1145/3583131.3590379,10542726} and classification models robustness testing \cite{bouchoucha2023robustness}, these approaches have not been extended to image segmentation. Segmentation models operate on high-dimensional, pixel-level outputs and require strict preservation of both visual and semantic integrity—challenges that are unique to this domain. Our approach, therefore, represents the first application of metamorphic testing with GA to assess and enhance segmentation robustness. This novel application not only tailors the optimization process to address the specific challenges of segmentation but also demonstrates superior performance against traditional gradient-based adversarial attacks.
Our first finding shows that traditional gradient-based adversarial attacks decrease the mean Intersection over Union (mIoU), a metric that quantifies segmentation accuracy by measuring overlap between predicted and ground truth regions, to a minimum of 8.5\% at a PSNR of 21.8 dB. In comparison, SegRMT can decrease the mIoU even further to a minimum of 6.4\% at a higher PSNR of 24 dB. This indicates that SegRMT often generates more challenging adversarial examples than traditional gradient-based attacks. The second finding demonstrates that SegRMT enhances the robustness of models fine-tuned by self-adversarial testing, but it may not surpass traditional gradient-based adversarial attacks. The third finding shows that the model fine-tuned on SegRMT adversarial achieves a maximum of 68.0\% as mIoU against the traditional gradient-based adversarial examples, while other models fine-tuned on traditional gradient-based adversarial achieve a maximum of 10.0\% as mIoU against SegRMT adversarial examples.

The remainder of the paper is organized as follows. Section~\ref{BgRw} presents the related work on segmentation, robustness assessment, and metamorphic testing. Section~\ref{Pf} formulates the problem and identifies our research's objectives, constraints, and variables. Section~\ref{meth} describes the proposed approach, including the design and implementation of the metamorphic testing and adversarial training methods. Section~\ref{Exp} illustrates our experimental setup and reports our findings on evaluating and enhancing the robustness of segmentation models. Finally, Section~\ref{conc} concludes the paper and discusses potential future work.

\section{Background and related work} \label{BgRw}
Recently, assessing the robustness of deep learning models, particularly for complex data types like RGB images, has become a critical area of research. In this section, we provide an overview of the existing literature, focusing on evaluating and enhancing the robustness of image segmentation models. This includes adversarial attacks, metamorphic testing, and hybrid approaches.

\subsection{RGB image segmentation}

Image segmentation is crucial for computer vision applications such as autonomous driving, medical imaging, and object recognition, all of which heavily rely on RGB images \cite{riehle2020robust}. This process involves effectively analyzing the red, green, and blue data channels in an image to extract essential details while dealing with the surroundings ~\cite{minaee2021image}. For instance, in autonomous driving scenarios, these RGB channels work together to help distinguish between various road elements - the red channel might highlight traffic signs, while the combination of channels helps differentiate between vehicles and road surfaces, even when they share similar gray tones.
By implementing deep learning models \cite{janiesch2021machine}, a significant reduction in time consumption and inaccurate predictions has been achieved in image segmentation \cite{minaee2021image}. However, these improvements come with new challenges, particularly in ensuring model robustness against adversarial data in real-world applications~\cite{akhtar2018threat}.
The fundamental challenges in RGB image segmentation manifest in several ways. In uncontrollable or unpredictable surroundings, models often struggle with accurate segmentation due to environmental variations \cite{kimai}. For example, a model that performs well in clear daylight might struggle during dawn or dusk, when shadows and reduced visibility alter the appearance of objects.
This difficulty is compounded by two key phenomena: low inter-class variance and high intra-class variance \cite{venkataramanan2021tacklinginterclasssimilarityintraclass}. Low inter-class variance occurs when different classes share similar visual traits, making it challenging for models to distinguish between them, particularly under varying conditions of perspective, lighting, and occlusions. Consider an urban scene where gray vehicles might appear similar to concrete buildings under certain lighting conditions, or where the boundary between a sidewalk and road becomes less distinct during rainy conditions.
Conversely, high intra-class variance presents itself when instances of the same class appear differently due to these same environmental factors, adding another layer of complexity to the segmentation task.  For instance, a single class like 'vehicle' can vary dramatically in appearance depending on the make, model, viewing angle, and lighting conditions - a red sports car viewed from the front may look entirely different from a white SUV seen from the side, yet both belong to the same class.
These challenges become particularly critical when considering model robustness . The interaction between environmental variations and class variance issues can significantly impact segmentation accuracy, especially in safety-critical applications \cite{e24091204}. Therefore, it is essential to evaluate and enhance the robustness of these models so that they can operate accurately and consistently across a wide range of real-world scenarios ~\cite{hendrycks2019b}.

\subsection{Evaluating robustness using adversarial attacks}

Some studies suggest that conducting adversarial attacks on segmentation models can offer valuable insights into their robustness. The purpose is to assess the model's performance against such attacks over time~\cite{hendrycks2019b, akhtar2018threat,croce2022evaluating}. Usually, these attacks apply tiny and non-random perturbations to the input data, leading the model to erroneous predictions. An early study proposes the concept of adversarial instances, which are dataset samples that the models often classify incorrectly. These data are studied to extract their features and then applied to a few other instances in the original dataset to examine their impact on the model's performance~\cite{goodfellow2015}. This provides the foundation for other attacks, such as gradient-based adversarial attacks.

The Fast Gradient Sign Method (FGSM) is one of the most straightforward and widely used adversarial attacks. It involves perturbing input data, such as pixels in images, by adding slight noise derived from the gradient of the loss with respect to the input~\cite{goodfellow2015,upadhyay2016fast,gu2022segpgd}.

Projected Gradient Descent (PGD) represents an advanced iterative approach derived from FGSM. It applies small, iteratively adjusted perturbations to the input data while maximizing the model's prediction error. Although it requires more time than FGSM, it is regarded as a more powerful attack~\cite{madry2017towards,gu2022segpgd}.

Carlini \& Wagner (C\&W) Attack is a renowned and effective attack that formulates the adversarial attack as an optimization problem. It minimizes the perturbation required to change the model's prediction while maintaining the adversarial example within a specified norm constraint. This method is highly effective but computationally intensive~\cite{carlini2017towards,klingner2020improved}. \newline
While non-gradient approaches have shown success in image classification tasks, gradient-based methods like FGSM, PGD, and C\&W remain the most effective for semantic segmentation. This is primarily due to the dense prediction nature of segmentation tasks, where each pixel requires individual consideration. Recent studies by Xie et al.~\cite{xie2017adversarialexamplessemanticsegmentation} and Arnab et al.~\cite{arnab2018robustnesssemanticsegmentationmodels} demonstrate that gradient-based methods are particularly well-suited for computing such pixel-wise perturbations. Attempts to adapt non-gradient methods to segmentation tasks have faced significant challenges due to the high-dimensional output space and spatial regularization requirements. For instance, even GAN-based approaches like AdvSPADE~\cite{shen2019advspaderealisticunrestrictedattacks} show inferior performance compared to gradient-based methods without substantial architectural modifications. Therefore, our evaluation focuses on these established and proven gradient-based approaches as they represent the current state-of-the-art in semantic segmentation attacks.

\subsection{Metamorphic testing and PSNR}

Fundamentally, metamorphic testing addresses the test case Oracle problem by creating new test cases based on successful existing ones~ \cite{chen2018metamorphic}. Therefore, using MT for segmentation model validation broadens the scope of evaluation beyond the basic testing dataset. The core principle of Metamorphic Testing (MT) lies in the identification and establishment of metamorphic relations (MRs), which outline the expected changes in output when specific modifications are made to the input. In deep learning, MRs are instrumental in testing models by examining how minor adjustments to the input, such as scaling or rotating an image, impact the output. Any divergence from the anticipated MR may indicate a potential issue with the model~\cite{xie2009application}. This also provides the opportunity to enhance models performance and generalizability before deploying for real-world applications.

Several studies leverage MRs to gain deep insights into the robustness and generalizability of image classification models by employing diverse image transformations~\cite{xie2009application,zhou2018metamorphic}. These transformations, encompassing rotation, scaling, and adjustments to color space, are devised to replicate common image distortions. Findings show that models validated using transformed images are often more robust against unseen test datasets. Other studies also indicate that incorporating MT allows developers to build more accurate and robust image classification models, particularly for high-stakes applications such as healthcare and autonomous navigation systems~\cite{chen2018metamorphic,xu2018enhancing,zhang2021deepbackground}.

Preserving the semantic integrity of images is essential for accurate robustness assessment, as excessive alteration can compromise the data. Using the Peak Signal-to-Noise Ratio (PSNR) as a metric can effectively measure the image integrity following pixel perturbations~\cite{huynh2008scope}. The higher the PSNR value, the lower the level of image corruption, but it does not necessarily indicate a low level of distortion. For example, an image with a PSNR value $>$ 20 dB is often regarded as non-corrupted and of sufficient quality for visual recognition; this is a pre-defined PSNR threshold in previous studies~\cite{1556624,bouchoucha2023robustness}. Therefore, employing the PSNR threshold during robustness testing ensures legitimate adversaries that contain valid data without severe corruption, which leads to accurate evaluation~\cite{wang2004image,ding2017metamorphic}.

Although the Structural Similarity Index Measure (SSIM) is widely recognized for its ability to capture perceptual quality by assessing structural similarity between images \cite{1284395}, its integration into our framework poses several challenges. SSIM is computationally more intensive than PSNR and is sensitive to slight variations, which may complicate its use as a hard threshold within our optimization process. In contrast, PSNR’s simplicity and the well-established 20 dB threshold \cite{bouchoucha2023robustness} provide a clear, objective criterion for filtering out over-distorted images. Given the high-dimensional nature of segmentation tasks—where pixel-level accuracy and preservation of semantic content are critical—the straightforward computation and interpretability of PSNR make it a more practical choice   \cite{5596999}. This allows our framework to consistently balance adversarial perturbation with image fidelity during the generation of diverse adversarial datasets.

In this work, we propose an approach that includes metamorphic relations and parameter constraints to generate realistic perturbations to test the models with various non-corrupted adversaries. This approach generates several diverse adversarial datasets from the original dataset, facilitating multiple model testing and adversarial training runs.

\section{Problem formulation} \label{Pf}

In this section, we define and structure the research problem. Also, we identify the objectives, constraints, problem specifications, and variables involved in our research.

In this study, we rigorously investigate the robustness of segmentation models tailored to diverse imaging modalities, encompassing hyperspectral, multispectral, and standard RGB datasets. Our dataset, \( D \), is formally defined as: \( D = \{(X_i, Y_i)\}_{i=1}^{N} \), where each \( X_i \) represents an image instance from the aforementioned modalities and \( Y_i \) denotes its corresponding accurate ground truth segmentation map. The primary objective for our segmentation model \( F \) is the precise mapping of each image \( X_i \) to its expected segmentation output, aiming for a high fidelity approximation \( F(X_i) \approx Y_i \).

To methodically assess and enhance the model’s resilience against various input perturbations, we employ an array of metamorphic relations \( R \).
These relations represent systematic ways to transform images while preserving their essential characteristics. For instance, in an autonomous driving context, if we slightly adjust an image's brightness (simulating different times of day) or add minor noise (simulating sensor interference), a car should still be recognized as a car in the segmentation output. 
Each metamorphic relation \( r \in R \) defines a deliberate, parameterized transformation \( T_r \), conceived to simulate potential real-world alterations affecting the images: \( T_r(X_i, \theta_r) = \hat{X}_i \), where \( \theta_r \) encapsulates the parameters of the transformation, constrained within a defined permissible range \( \Theta \).

The core premise of our metamorphic testing protocol insists that, despite these transformations, the essential semantic integrity of the image segments must be preserved, i.e., \( F(T_r(X_i, \theta_r)) \approx Y_i \). This condition forms the basis for asserting the robustness of our model, ensuring that the semantic content of the segments remains intact despite the application of \( T_r \).

We approach this challenge by formulating a constrained optimization problem designed to maximize a robustness criterion \( C(\hat{X}) \), which quantitatively evaluates the model’s ability to uphold segmentation accuracy in the face of these synthetic perturbations. This is mathematically expressed as:
\[
\text{Maximize } C(\hat{X}) \text{ subject to } \Phi_i(\hat{X}) = 0, \text{ for } i = 1, \ldots, u
\]
Here, \( \Phi \) represents a set of validity constraints ensuring that the transformations \( T_r \) respect the semantic boundaries as defined by the original ground truths. Specifically, our validity constraints consist of a Peak Signal-to-Noise Ratio (PSNR) threshold of 20dB and realistic transformation requirements. The PSNR threshold ensures that our transformations maintain sufficient image quality and visual coherence while still allowing for meaningful perturbations. The realism constraints guarantee that our transformations simulate real-world scenarios that could naturally occur during image acquisition and processing, such as lighting variations, sensor noise, or perspective changes. Together, these constraints ensure that our adversarial examples remain both challenging and representative of real-world conditions that a segmentation model might encounter during deployment.

To augment the model’s resilience further, adversarially generated examples \( D_{\text{adv}} \) — synthetic inputs that significantly deviate from expected outcomes under nominal conditions — are integrated into the training dataset \( D_{\text{train}} \):
\[
D_{\text{augm}} = D_{\text{train}} \cup D_{\text{adv}}
\]
Subsequently, the model \( F \) undergoes fine-tuning on \( D_{\text{augm}} \) aimed at minimizing the empirical error, measured via a loss function \( L \), across both original and adversarial examples:
\[
\text{Minimize } \mathbb{E}[(X, Y) \sim D_{\text{augm}}] [L(F(X), Y)]
\]
This meticulous approach, employing metamorphic testing and adversarial training, not only fortifies the segmentation models against a spectrum of challenging conditions but also systematically refines their accuracy and reliability across varied imaging contexts. This methodology ensures a comprehensive evaluation and continuous enhancement of the models’ segmentation capabilities, embodying a robust defense against real-world perturbations and synthetic adversarial tactics.

\section{Methodology}\label{meth}

In this section, we describe the use of metamorphic testing combined with the PSNR constraints. Also, we explain how we optimize image perturbations using the genetic algorithm.

We employ a metamorphic testing framework to evaluate our segmentation model's robustness thoroughly and rigorously. This framework systematically applies a series of controlled transformations to the input images, known as metamorphic relations. These transformations are designed to simulate a variety of real-world perturbations and variations, thereby providing a comprehensive assessment of the model's performance under challenging conditions.

\begin{figure}[h]
    \centering
     \includegraphics[width=1.02\linewidth]{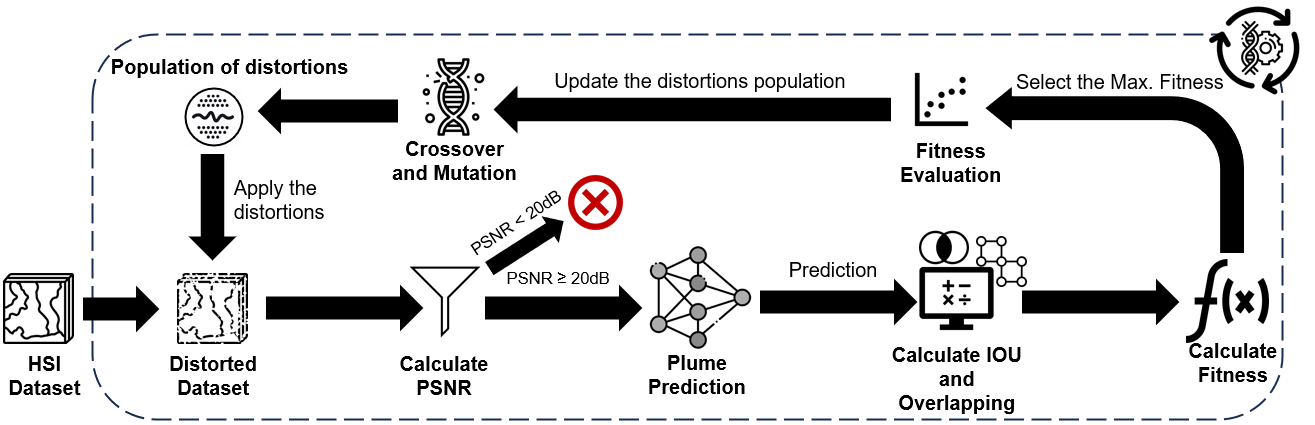}
    \caption{Pipeline of the proposed SegRMT for robustness assessment. The pipeline illustrates the process from initial image perturbation using various transformations and optimization using the genetic algorithm to the evaluation of segmentation model performance.}
    \label{fig:pipeline}
\end{figure}

\begin{figure}[h!]
    \centering
    \includegraphics[width=\linewidth]{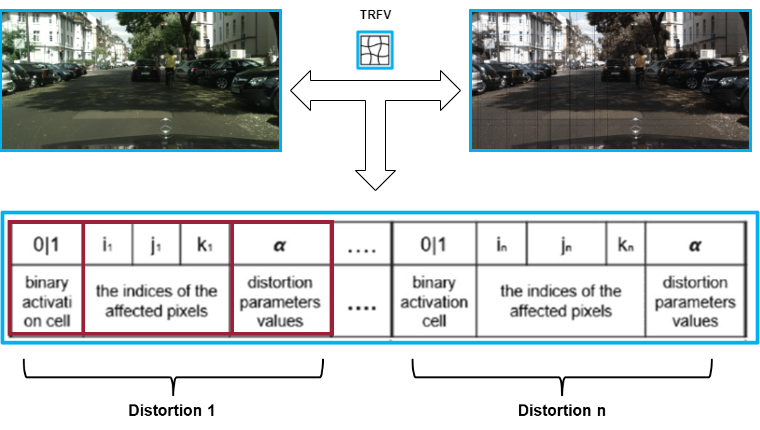} 
    \caption{Transformation vector structure.}
    \label{fig:structure}
\end{figure}

Figure~\ref{fig:pipeline} shows the overall process of generating and evaluating transformation vectors, which are crucial in SegRMT. The process consists of two main components highlighted in Figure 1. The first component is the transformation vector generation, where vectors are randomly initialized. Each vector consists of sub-transformation vectors representing specific types of noise or perturbation. As shown in Figure 2, each sub-transformation vector includes 1) a binary activation cell indicating whether the transformation is active, 2) indices of the affected pixels, and 3) distortion parameter values. The second component is the genetic algorithm optimization, which evaluates and evolves these transformation vectors to find the most effective perturbations. The GA employs a carefully designed fitness function that balances two objectives: maximizing the PSNR to ensure image quality while minimizing the Intersection over Union (IoU) to identify transformations that significantly impact model performance. Through this optimization process, the GA aims to discover the least perceptible transformations that can effectively challenge the segmentation model.

\subsection{Image Transformation}
In real-world deployments, image segmentation models must handle various forms of image degradation. Our transformation selection methodology addresses these challenges through two fundamental categories: spatial and spectral distortions, reflecting the primary ways images can be compromised in practical applications.
This categorization is motivated by both empirical studies and practical considerations. For example, Hendrycks et al. \cite{hendrycks2019benchmarkingneuralnetworkrobustness}demonstrated that a wide range of noise, blur, and digital corruptions—similar to those we consider—can significantly affect model performance, while Taori et al. \cite{10.5555/3495724.3497285} highlighted the importance of testing models against natural distribution shifts using diverse perturbations.
Spatial distortions simulate physical and geometric alterations to the image structure. We focus on four key types: region dropout for occlusions and missing data, line/column transformations for systematic sensor errors, salt and pepper noise for impulse distortions, and Gaussian noise for thermal variations. Spectral distortions, in contrast, target color information processing through channel dropout and channel-specific noise, simulating color degradation scenarios common in real imaging systems.

\subsubsection{Spatial Distortions}

Let $I: \Omega \to \mathbb{R}^C$ be an image defined on the pixel grid $\Omega \subset \mathbb{Z}^2$, where $C$ is the number of channels (e.g., $C=3$ for RGB images). For clarity, we define:
\begin{itemize}
    \item $\text{MIN}_I = \min_{(x,y) \in \Omega} I(x,y)$,
    \item $\text{MAX}_I = \max_{(x,y) \in \Omega} I(x,y)$,
    \item $\text{CONST}_I$: a predetermined constant value (typically chosen as either $\text{MIN}_I$ or $\text{MAX}_I$).
\end{itemize}

Spatial distortions are applied to test the model's ability to handle changes in the spatial domain. These distortions include:

\begin{itemize}
    \item \textbf{Region Dropout:}
    \begin{itemize}
        \item \textbf{Description:} Simulates occlusions or missing data by randomly altering regions in the image.
        \item \textbf{Purpose:} This transformation mimics real-world scenarios where parts of an image might be blocked by objects, sensor malfunctions, or environmental obstructions. In segmentation tasks, occlusions are common—such as pedestrians partially hidden behind vehicles or objects obscured by shadows. By applying region dropout, the model is forced to rely on contextual cues from the remaining visible parts, thereby testing its ability to infer and preserve semantic information even when significant regions are missing.
        \item \textbf{Mathematical Formulation:} For each pixel $(x,y) \in \Omega$, define:
        \[
        I'(x, y) = 
        \begin{cases} 
        \text{MIN}_I, & \text{with probability } p_{\text{min}}, \\[1mm]
        I(x, y), & \text{with probability } p_{\text{unchanged}}, \\[1mm]
        \text{MAX}_I, & \text{with probability } p_{\text{max}},
        \end{cases}
        \]
        where $p_{\text{min}} + p_{\text{unchanged}} + p_{\text{max}} = 1$.
    \end{itemize}
    
    \item \textbf{Line and Column Transformations:}
    \begin{itemize}
        \item \textbf{Description:} Modifies entire rows or columns of pixels to simulate sensor errors or calibration issues. Such transformations have been shown to effectively model structured perturbations \cite{}.
        \item \textbf{Purpose:} These transformations represent systematic, structured distortions that can occur due to hardware issues (e.g., faulty sensor lines) or calibration errors in imaging devices. Such errors can lead to consistent distortions across an image. For segmentation, where continuity and precise boundaries are critical, these structured perturbations test the model’s resilience against uniform or patterned noise, ensuring that it can still accurately delineate object boundaries despite consistent, directionally biased distortions.
        \item \textbf{Continuous Line/Column Dropout:} Let $\ell$ denote a specific row or column index. Then, for all $(x,y) \in \Omega$, one can set:
        \[
        I'(\ell, y) = \text{CONST}_I \quad \text{or} \quad I'(x, \ell) = \text{CONST}_I.
        \]
        \item \textbf{Line Stripping:} For a given stride $s \in \mathbb{N}$, define:
        \[
        I'(x, y) = 
        \begin{cases} 
        I(x, y), & \text{if } x \bmod s \neq 0, \\[1mm]
        \text{CONST}_I, & \text{if } x \bmod s = 0,
        \end{cases}
        \]
        where $x \bmod s$ denotes the remainder when $x$ is divided by $s$. A similar formulation applies for column stripping.
    \end{itemize}
    
    \item \textbf{Salt and Pepper Noise:}
    \begin{itemize}
        \item \textbf{Description:} Introduces random occurrences of black and white pixels.
        \item \textbf{Purpose:} Salt and pepper noise is a classic model for impulse noise, often arising from errors in data transmission or sensor defects. In practical imaging scenarios, sudden and isolated pixel-level disturbances may occur due to environmental interference or hardware glitches. For segmentation models, handling such abrupt changes without losing overall structural information is crucial. This transformation challenges the model to remain robust in the presence of isolated, high-contrast pixel anomalies.
        \item \textbf{Mathematical Formulation:} For each pixel $(x,y) \in \Omega$,
        \[
        I'(x, y) = 
        \begin{cases} 
        \text{MIN}_I, & \text{with probability } p_{\text{salt}}, \\[1mm]
        I(x, y), & \text{with probability } 1 - \big(p_{\text{salt}} + p_{\text{pepper}}\big), \\[1mm]
        \text{MAX}_I, & \text{with probability } p_{\text{pepper}},
        \end{cases}
        \]
        with $p_{\text{salt}} + p_{\text{pepper}} \leq 1$.
    \end{itemize}
    
    \item \textbf{Spatial Gaussian Noise:}
    \begin{itemize}
        \item \textbf{Description:} Adds Gaussian-distributed noise to the pixel values.
        \item \textbf{Purpose:} Gaussian noise represents natural fluctuations that occur in sensor readings (e.g., thermal noise). Unlike impulse noise, Gaussian noise is spread throughout the image and tends to be less abrupt, but it still affects the clarity of edges and textures. By introducing spatial Gaussian noise, the model is tested on its ability to distinguish important structural features from the inherent noise present in real-world imaging, ensuring that minor variations in pixel intensities do not lead to significant mis-segmentation.
        \item \textbf{Mathematical Formulation:}
        \[
        I'(x, y) = I(x, y) + \eta(x,y),
        \]
        where $\eta(x,y) \sim N(\mu, \sigma^2)$ are independent samples drawn from a Gaussian distribution with mean $\mu$ and variance $\sigma^2$, for each $(x,y) \in \Omega$.
    \end{itemize}
\end{itemize}

\subsubsection{Spectral Distortions}
Spectral distortions test the model's robustness to variations in color channels. Although these are typically applied to multi-spectral images, they can be adapted to RGB images.

\begin{itemize}
    \item \textbf{Channel Dropout:}
    \begin{itemize}
        \item \textbf{Description:} Simulates the loss of specific color channels in RGB images.
        \item \textbf{Purpose:} In real-world conditions, sensors may sometimes fail to capture complete color information due to hardware faults or adverse lighting conditions. Channel dropout forces the segmentation model to operate with incomplete color information, testing its robustness to missing data. This is especially critical for segmentation tasks where color cues often play a significant role in differentiating between objects with similar shapes but different colors.
        \item \textbf{Mathematical Formulation:}
        \[
        I'_c(x, y) = \text{CONST}_I,
        \]
        where $c \in \{R, G, B\}$ denotes a specific color channel.
    \end{itemize}
    
    \item \textbf{Channel Gaussian Noise:}
    \begin{itemize}
        \item \textbf{Description:} Introduces Gaussian noise to specific color channels in RGB images.
        \item \textbf{Purpose:}This transformation addresses situations where one or more color channels might exhibit slight, channel-specific variations due to environmental changes or sensor inconsistencies. Since color fidelity is important for accurately distinguishing between objects—especially in scenarios where objects have similar textures but differing color profiles—this transformation ensures that the model can handle slight fluctuations in individual channels without degrading segmentation performance. It also reinforces the idea that the model should not be overly sensitive to minor color variations, which are common in real-world settings.\cite{knauthe2024transparencydistortionrobustnesssota}.
        \item \textbf{Mathematical Formulation:}
        \[
        I'_c(x, y) = I_c(x, y) + \eta_c(x,y),
        \]
        where $\eta_c(x,y) \sim N(\mu, \sigma^2)$ represents the noise added to channel $c$.
    \end{itemize}
\end{itemize}

Together, these transformations capture many common real-world degradations. We selected region dropout, line/column transformations, salt and pepper noise, and spatial Gaussian noise for the spatial domain because they effectively simulate issues such as occlusions, sensor malfunctions, and natural noise variations—conditions frequently encountered in applications like autonomous driving. Similarly, the spectral perturbations (channel dropout and channel-specific Gaussian noise) mimic failures in color capture due to hardware faults or adverse lighting conditions. These choices are grounded in established research \cite{10.1007/978-3-031-78201-5_12,hendrycks2019benchmarkingneuralnetworkrobustness,10.5555/3495724.3497285,knauthe2024transparencydistortionrobustnesssota}, ensuring that our transformation set is both comprehensive and relevant to practical scenarios. While this set robustly challenges segmentation models, future work may explore additional transformations (e.g., geometric rotations, scaling, and weather-induced effects) to further expand the evaluation of model robustness.

\subsection{Robustness Criterion}
Let $F(\hat{X}_i)$ denote the segmentation output for the perturbed image $\hat{X}_i$, and let $Y_i$ be the corresponding ground truth segmentation map. The robustness criterion, which quantifies the model's ability to maintain segmentation accuracy under perturbations, is defined as:
\[
C(\hat{X}) = \frac{1}{N} \sum_{i=1}^{N} \text{IoU}\big(F(\hat{X}_i), Y_i\big),
\]
where $\text{IoU}$ (Intersection over Union) measures the overlap between the predicted segmentation $F(\hat{X}_i)$ and the ground truth $Y_i$. A higher $C(\hat{X})$ value indicates better robustness.

\subsection{Genetic Algorithm for Optimizing Transformations}

The pursuit of reliable segmentation models led to the identification of a notable gap in current approaches related to the generation of adversarial examples. Although fixed transformations and random perturbations have their advantages, adopting a more advanced methodology has the potential to provide adversarial examples that are both more realistic and sophisticated. In light of this revelation, the investigation of Genetic Algorithms (GAs) in this particular field was initiated.

Genetic Algorithms (GAs), which are based on principles from evolutionary biology, provide a promising foundation for effectively navigating the intricate search space of image transformations. The underlying logic was that by emulating the process of natural selection, it could be feasible to develop progressively more efficient combinations of transformations, thus expanding the limits of what conventional approaches could achieve.
Following a comprehensive examination of relevant academic articles and careful consideration, the determination was reached to employ a customized Genetic Algorithm (GA) specifically designed to address the unique requirements of this study. The choice to enhance the complexity of the project was considered appropriate due to the potential benefits it could bring in terms of improving the quality and diversity of adversarial examples.

The design of the chromosome structure is a critical component of the genetic algorithm, directly influencing its ability to represent and evolve effective transformations. After careful consideration and multiple iterations, a sophisticated chromosome structure was developed to encode complex sequences of transformations.

Each chromosome consists of several sub-transformation vectors, each representing a specific type of distortion that can be applied to the input image. Figure~\ref{fig:structure} shows the structure of each sub-transformation vector is as follows:

\begin{itemize}
    \item \textbf{Binary Activation Cell}: A single bit (0 or 1) indicating whether this particular distortion should be applied.
    \item \textbf{Distortion Parameters}: A set of values specific to the type of distortion (e.g., dropout rate, noise variance, color shift values).
    \item \textbf{Affected Indices}: Specifies which pixels or bands the distortion should be applied to.
\end{itemize}

This detailed encoding ensures a deterministic mapping between the chromosome and the resulting distorted input, given the original image. It allows for fine-grained control over the application of distortions while maintaining the flexibility to represent a wide variety of transformation combinations.

The population initialization process was carefully designed to generate a diverse set of valid chromosomes. This involved:
\begin{enumerate}
    \item Randomly determining the number of sub-transformation vectors for each chromosome.
    \item For each sub-transformation vector:
    \begin{enumerate}
        \item Randomly setting the activation bit.
        \item Generating appropriate parameter values within predefined ranges specific to each distortion type.
        \item Selecting affected indices based on the distortion type and image dimensions.
    \end{enumerate}
\end{enumerate}

A validation step was implemented to ensure that all initial chromosomes represented feasible transformation sequences. This extra layer of validation significantly reduced errors in subsequent generations and ensured that the genetic algorithm started with a population of viable solutions.

This chromosome design, coupled with the carefully crafted initialization process, provided a solid foundation for the genetic algorithm to explore and evolve increasingly effective combinations of image transformations.

A key innovation in our GA implementation is the development of a sophisticated fitness function that balances two competing objectives: maximizing the disruption of segmentation results while preserving image fidelity. After extensive experimentation, we developed the following formulation:
\[
F = 
\begin{cases} 
(1 - \text{IoU}) \times \left(\frac{\text{PSNR}}{20}\right) & \text{if PSNR} \geq 20 \text{ dB} \\
0 & \text{if PSNR} < 20 \text{ dB}
\end{cases}
\]

This formulation incorporates several key insights gained through the research process:

\begin{itemize}
    \item \textbf{PSNR Normalization}: By dividing PSNR by the threshold value of 20 dB, the function creates a balanced interplay between segmentation disruption (measured by IoU) and image fidelity (measured by PSNR). This normalization ensures that neither objective dominates the fitness calculation.
    \item \textbf{IoU Inversion}: The use of $(1 - \text{IoU})$ in the formula ensures that lower IoU values, which indicate greater segmentation disruption, result in higher fitness scores. This aligns the fitness function with the goal of finding transformations that significantly impact segmentation performance.
    \item \textbf{Quality Threshold}: The implementation of a hard cutoff at 20 dB PSNR serves to eliminate transformations that excessively degrade image quality. This threshold was determined through a combination of literature review \cite{bouchoucha2023robustness} and empirical testing, representing a balance point between perceptible image degradation and effective adversarial perturbation.
\end{itemize}

Through comprehensive testing across diverse transformation scenarios, this fitness function has demonstrated consistent ability to guide the GA towards transformations that are both subtle in visual impact and effective in disrupting segmentation performance. The combination of our sophisticated chromosome structure and carefully crafted fitness function provides a solid foundation for the GA to explore and evolve increasingly effective combinations of image transformations.

\subsection{Genetic Operators and Algorithm Configuration \newline}
The design of our genetic algorithm required three key genetic operators to effectively explore the transformation space. The first operator is a tournament selection mechanism that selects chromosomes based on their fitness values. This selection process adapts based on the current population's diversity to prevent premature convergence and maintain exploration throughout all generations.
For the crossover operation, we developed a two-point crossover specifically for our variable-length chromosome structure. Since each chromosome represents a sequence of image transformations, the crossover points must occur between complete transformations to maintain valid sequences. This ensures that when genetic material is exchanged between parent chromosomes, the resulting offspring contain properly structured transformation sequences.
The third operator is a mutation mechanism that works at three different levels. At the first level, it can add new transformations to the sequence or remove existing ones. At the second level, it can change the type of a transformation, for example changing a brightness adjustment to a contrast adjustment. At the third level, it can modify the specific parameters within a transformation, such as adjusting the intensity of a noise filter.
The mutation operator is particularly important for maintaining genetic diversity and exploring new transformation combinations. By operating at multiple levels, it allows for both large changes in the transformation sequence and fine-tuning of specific parameters. This helps the algorithm explore different combinations of transformations while also optimizing their parameters.
This genetic operator design focuses on preserving valid transformation sequences while allowing sufficient exploration of the transformation space to find effective adversarial examples.

\subsection{Integration with Metamorphic Testing}
Our framework integrates genetic algorithms with metamorphic testing by establishing a systematic relationship between transformation sequences and genetic operations. At the core of this integration is a transformation engine that processes both spatial and spectral distortions according to defined metamorphic relations. The engine employs a bidirectional translation mechanism between chromosome representations and metamorphic transformations, ensuring that genetic operations maintain semantic meaning in the context of image perturbation.
The integration follows a sequential transformation approach, where multiple distortions can be combined to create complex perturbation patterns. This sequential processing allows for the composition of transformations that better reflect real-world scenarios while maintaining the validity of metamorphic relations. Each transformation sequence is validated against both genetic constraints and metamorphic testing principles, ensuring that evolved sequences remain semantically meaningful.
Through this integration, metamorphic testing principles guide the genetic algorithm's exploration of the transformation space, while genetic operations enable systematic discovery of effective transformation combinations. This synergy between metamorphic testing and genetic optimization creates a framework that can effectively generate and evaluate adversarial examples while maintaining the semantic integrity of the transformations.

\subsection{Preserving Semantic Integrity}
To ensure the robustness of our segmentation models while preserving the semantic integrity of the images, we employ two primary techniques: setting a Peak Signal-to-Noise Ratio (PSNR) threshold and defining parameter constraints for noise transformations.

The Peak Signal-to-Noise Ratio (PSNR) is used as a quality metric to preserve the semantic integrity of the images. PSNR measures the ratio between the maximum possible power of a signal and the power of corrupting noise, with higher values indicating better quality.

In our approach, we set a PSNR threshold of 20 dB. According to the literature, a PSNR of 20 dB or higher typically preserves the essential content of the image, preventing it from being perceived as overly corrupted \cite{bouchoucha2023robustness}. Images with a PSNR greater than 20 dB are retained, while those with a PSNR of 20 dB or lower are discarded. This ensures that the introduced noise does not significantly distort the images, thereby maintaining their semantic content.

By applying this PSNR threshold, we ensure that the images used for evaluating model robustness maintain their structural and semantic integrity, providing a realistic yet challenging test for the segmentation models.

We define the maximum allowable percentage of pixels that can be affected by various noise transformations through a parameter file. This parameter file was meticulously fine-tuned through an iterative process to achieve the optimal balance between introducing perturbations and preserving semantic integrity.

\paragraph{Fine-Tuning Process:}

\begin{enumerate}
    \item \textbf{Initial Parameter Range:} We began by establishing a broad range for each parameter, encompassing the maximum and minimum possible values for the percentage of affected pixels, noise levels, and transformation extents. This initial range allowed us to explore the full spectrum of potential transformations.
    \item \textbf{Iterative Testing and Evaluation:} We then applied these transformations to a subset of images and evaluated their impact on semantic integrity using a combination of visual inspections and quantitative metrics. The evaluation focused on whether the transformed images retained their essential semantic features and whether the segmentation model's performance remained robust.
    \item \textbf{Narrowing the Range:} Based on the evaluation results, we progressively narrowed the parameter ranges. Parameters that resulted in excessive degradation or loss of semantic content were adjusted to reduce their impact. This iterative refinement continued until we identified parameter values that consistently preserved semantic integrity while still challenging the segmentation model.
    \item \textbf{Optimization for Sweet Spot:} The final parameter values represent a "sweet spot" where the transformations introduce sufficient perturbations to test the model's robustness without compromising the semantic content of the images. This balance ensures that the transformations simulate realistic variations and noise conditions while maintaining the integrity of the underlying semantic structures.
\end{enumerate}

\subsection{Adversarial Training }
Our adversarial training framework systematically evaluates different approaches to enhancing model robustness through adversarial examples. Rather than combining all types of adversarial examples into a single training set, we developed a structured approach that assesses the effectiveness of each method independently. The framework begins with the generation of adversarial examples using both traditional gradient-based methods (FGSM, PGD, C\&W) and our proposed SegRMT approach. For all generated examples, we maintain a PSNR threshold above 20 dB to ensure semantic integrity is preserved.
For training dataset construction, we create separate configurations, each combining clean data with adversarial examples from a single method. This separation allows us to precisely evaluate how each type of adversarial training affects model performance. We maintain consistent ratios between clean and adversarial examples across all configurations to ensure fair comparison, while applying standard augmentation techniques uniformly across all datasets.
The training strategy involves training separate models using each adversarial dataset configuration. Each model employs a composite loss function that combines standard cross-entropy with specific penalties for adversarial examples, encouraging robust feature learning. Throughout training, we conduct regular validation on both clean and adversarial validation sets to monitor the model's progress in both standard accuracy and adversarial robustness.
Our cross-evaluation protocol tests each trained model against all types of adversarial examples, not just those it was trained on. This comprehensive evaluation reveals how well the robustness gained from one type of adversarial training generalizes to other types of attacks. Additionally, we carefully measure any performance degradation on clean data to ensure that improved robustness does not come at too great a cost to standard performance. This systematic approach enables us to thoroughly analyze and compare how different types of adversarial training affect model robustness, providing insights into which methods offer the best balance of specific and general robustness improvements.

\section{Experiments} \label{Exp}
In this section, we will detail the Experimental setup used to conduct this study we will also detail and discuss the different results obtained offering insights into their significance and implications
\subsection{Experimental Setup}

\textbf{Datasets:} We used the Cityscapes dataset for our experiments, which consists of 5,000 high-resolution urban street images with a resolution of 2048x1024 pixels. The dataset is divided into 2,975 training images, 500 validation images, and 1,525 testing images. This dataset was selected due to its complexity and relevance to autonomous driving applications, providing a challenging environment for testing model robustness.

\textbf{Models:}
Our experiments utilized the DeepLabV3 model with a ResNet-50 backbone, a well-established architecture for segmentation tasks. The model was pre-trained on the Cityscapes dataset, serving as a strong baseline for evaluating the impact of adversarial perturbations.

\textbf{Baseline Methods:}
 To generate adversarial examples, we implemented three widely recognized methods:
\begin{itemize}
  \item \textbf{Fast Gradient Sign Method (FGSM):} Implemented with an epsilon value of 0.09 to achieve a PSNR value of approximately 20.
  \item \textbf{Projected Gradient Descent (PGD):} Conducted with 10, 40, and 100 iterations using alpha and epsilon values of 0.08 and 0.09, respectively.
  \item \textbf{Carlini \& Wagner (C\&W) Attack:} Executed as an optimization problem with an epsilon value of 0.15 and a learning rate of 1e-5 to minimize perturbation while maintaining adversarial efficacy and a PSNR abover 20.
\end{itemize}

\textbf{Evaluation Metrics:\newline}
The effectiveness of the adversarial attacks was measured using two key metrics:
\begin{itemize}
  \item \textbf{Intersection over Union (IoU):}This metric assesses the overlap between the predicted segmentation and the ground truth, indicating the accuracy of the segmentation.
  \item \textbf{Peak Signal-to-Noise Ratio (PSNR):}This metric quantifies the perceptual similarity between the original and adversarial images, ensuring that the adversarial examples maintain the semantic integrity of the images.
\end{itemize}

\textbf{Genetic Algorithm Configuration:} Through extensive experimentation, we established the following parameters for our genetic algorithm: a population size of 50 individuals, evolution limit of 100 generations, crossover rate of 0.8, and mutation rate of 0.2 with variable sub-rates for different mutation types. We preserved the top 2 individuals through elitism and implemented early termination when fitness improvement remained below 0.1\% for 15 consecutive generations. This configuration provided an effective balance between exploration and exploitation.

\textbf{Statistical Considerations:\newline}
To ensure the robustness of the results and enable statistical analysis, each experiment was repeated 10 times using different random seeds. The results from these multiple runs were used to perform statistical significance tests, such as the Wilcoxon signed-rank test, and calculate effect sizes (e.g., Cohen's d). This allowed for a more reliable evaluation of the differences between the SegRMT method and traditional gradient-based adversarial attacks.

\subsection{Evaluating Segmentation Robustness }

To assess our segmentation model's robustness, we conducted comparative experiments between our SegRMT approach and traditional adversarial methods (FGSM, PGD, and C\&W attacks). Each method was calibrated to maintain a PSNR above 20 dB, ensuring fair comparison while preserving essential image content and semantics. Our evaluation used the DeepLabV3 model trained on the Cityscapes dataset, focusing on both original and adversarially perturbed images.
The results, presented in Table 1, demonstrate SegRMT's superior effectiveness in generating challenging adversarial examples. Our approach achieved a significant reduction in model performance, lowering the mIoU to 6.4\% while maintaining a higher PSNR of 24.0 dB. In comparison, traditional methods showed less effectiveness: FGSM reduced mIoU to 11.3\% (PSNR 20.6 dB), PGD variants achieved between 8.5\% and 9.4\% (PSNR 21.8 dB), and C\&W reached 21.7\% (PSNR 21.3 dB). These results indicate that SegRMT generates more potent adversarial examples while better preserving image quality.
Statistical analysis of our results, based on 10 repeated experiments with different random seeds, confirmed the significance of these findings. The Wilcoxon signed-rank test and Cohen's d effect size calculations demonstrated that SegRMT's performance improvements over traditional methods were both statistically significant and practically meaningful. This comprehensive evaluation framework revealed that our approach provides a more practical assessment of model robustness, particularly in scenarios requiring realistic perturbation patterns.

\subsection{Experimental Setup for Adversarial Training}
This section details the specific experimental configurations and implementation details used in our adversarial training evaluations.

\paragraph{Implementation Configuration}
We conducted our experiments using the DeepLabV3 model with a ResNet-50 backbone on the Cityscapes dataset. Each experiment was performed using one of five distinct configurations:
\begin{enumerate}
    \item Base model trained exclusively on clean data (baseline)
    \item Model trained with FGSM adversarial augmentation
    \item Model trained with PGD adversarial augmentation
    \item Model trained with C\&W adversarial augmentation
    \item Model trained with SegRMT adversarial augmentation
\end{enumerate}

\paragraph{Training Parameters}
For all configurations, we employed a Stochastic Gradient Descent (SGD) optimizer with an initial learning rate of 0.001, momentum of 0.9, and weight decay of 0.0005. Given the high-resolution nature of Cityscapes images (2048x1024 pixels) and GPU memory constraints, we implemented a batch size of 2. All adversarial examples were generated while maintaining a minimum PSNR threshold of 20 dB to ensure data integrity.

\paragraph{Data Processing}
Our implementation included standard data augmentation techniques: random resizing, cropping to 512x1024 pixels, horizontal flipping, and photometric distortion. For adversarial configurations, we maintained separate datasets combining clean images with their respective adversarial examples. The ratio between clean and adversarial examples was determined through preliminary experiments to optimize robustness while maintaining clean data performance.

\paragraph{Training Protocol}
Each model configuration underwent training for 80,000 iterations on an NVIDIA A100 GPU. We implemented early stopping when validation performance showed no improvement over 5,000 consecutive iterations. The learning rate followed a polynomial decay schedule from 0.001 to 0.0001 over the first 40,000 iterations. Performance monitoring was conducted on both clean and adversarial validation sets at 1,000-iteration intervals.

\paragraph{Evaluation Procedure}
Models were evaluated on three distinct test sets:
\begin{itemize}
    \item Clean test data to establish baseline performance
    \item Test data with adversarial examples from their respective training method
    \item Test data with adversarial examples from all other methods to assess cross-adversarial robustness
\end{itemize}

\subsection{Results and Analysis \newline}

\subsubsection*{RQ1: How effective is SegRMT compared to traditional gradient-based adversarial attacks in deceiving a segmentation model?
 \newline}

\paragraph{\textbf{Motivation:}}
This research question is motivated by the necessity to thoroughly assess and improve the resilience of segmentation models, especially in scenarios where adversarial attacks might significantly hinder model performance. Conventional gradient-based adversarial techniques, including FGSM, PGD, and C\&W, have been extensively used to stress-test the model used. However, they may not comprehensively capture the wide range of possible adversarial scenarios. By incorporating Metamorphic Testing (MT) and Genetic Algorithms (GA), SegRMT aims to investigate a broader spectrum of adversarial examples, potentially leading to more effective results.

\paragraph{\textbf{Approach:}}
To investigate the effectiveness of SegRMT, we subjected the DeepLabV3 model, trained on the Cityscapes dataset, to a comprehensive evaluation against various adversarial attacks. The resilience of the model was measured using two main metrics: mean Intersection over Union (mIoU) for segmentation accuracy and Peak Signal-to-Noise Ratio (PSNR) for evaluating the subtlety of the applied disturbances.

SegRMT, which integrates a Genetic Algorithm within a Metamorphic Testing framework, was systematically compared to traditional gradient-based adversarial methods, including FGSM, PGD, and C\&W. The Genetic Algorithm (GA) in SegRMT optimizes perturbations by balancing the trade-off between minimizing mIoU and maximizing PSNR. A threshold of 20 dB is established to ensure that the adversarial examples remain perceptually realistic. This method enabled a meticulous evaluation of the effectiveness of SegRMT in generating challenging yet visually subtle adversarial examples.

\begin{table}[h]
\centering
\caption{Robustness Testing Results on Cityscapes with DeepLabV3}
\label{tab:robustness_results}
\begin{tabular}{lcc}
\toprule
\textbf{Testing Method} & \textbf{mIoU} (\%) & \textbf{PSNR (dB)} \\ \midrule
Original (No Perturbation) & 79.4 & Inf \\ 
FGSM & 11.3 & 20.6 \\ 
PGD10 & 9.4 & 21.8 \\ 
PGD40 & 8.5 & 21.8 \\ 
PGD100 & 9.1 & 21.8 \\ 
C\&W & 21.7 & 21.3 \\ 
SegRMT & \textbf{6.4} & \textbf{24.0} \\ \bottomrule
\end{tabular}
\end{table}

\paragraph{\textbf{Results:}}
The results of the robustness testing, as shown in Table \ref{tab:robustness_results}, demonstrate the effectiveness of the SegRMT approach compared to traditional gradient-based adversarial attacks on the DeepLabV3 model trained with the Cityscapes dataset. The model’s initial performance on the unaltered dataset revealed a high mean Intersection over Union (mIoU) of 79.4\%, highlighting its precision in optimal circumstances.

However, the model’s ability to withstand adversarial perturbations significantly decreased, particularly with the SegRMT approach. The FGSM attack decreased the mean Intersection over Union (mIoU) to 11.3\%, with a Peak Signal-to-Noise Ratio (PSNR) of 20.6 dB, underscoring the model’s vulnerability to even basic gradient-based attacks. The iterative PGD method showed varying degrees of impact, with mIoU values ranging from 9.4\% to 8.5\% as iterations increased, though effectiveness plateaued beyond 40 iterations, indicating a diminishing return on computational effort.

The C\&W attack, known for its accuracy, yielded a mean Intersection over Union (mIoU) of 21.7\% and a Peak Signal-to-Noise Ratio (PSNR) of 21.3 dB. Nevertheless, SegRMT showed the most notable results by achieving the lowest mIoU of 6.4\% while preserving the highest PSNR of 24 dB. This suggests that the use of a Genetic Algorithm in SegRMT’s Metamorphic Testing framework enables the generation of highly effective adversarial instances that are both subtle and significantly impactful. These findings emphasize SegRMT’s exceptional capacity to deteriorate model performance, making it a robust tool for evaluating the resilience of segmentation models against a broader range of adversarial scenarios.

To assess the statistical significance of the differences in performance between the metamorphic testing tool and the gradient-based adversarial attack methods, we conducted the Wilcoxon signed-rank test and calculated Cohen's d effect size.
The Wilcoxon signed-rank test is a non-parametric statistical test used to compare two related samples or repeated measurements on a single sample to assess whether their population mean ranks differ. We performed pairwise comparisons between the metamorphic testing tool and each of the gradient-based methods (PGD10, PGD40, PGD100, CW, and FGSM) using the image-wise IoU scores. The results are as follows:
\begin{itemize}
\item Wilcoxon Test for SegRMT vs. PGD10: Test Statistic: 48918.0, P-value: 0.0047
\item Wilcoxon Test for SegRMT vs. PGD40: Test Statistic: 38221.0, P-value: 2.099${e}^{-10}$
\item Wilcoxon Test for SegRMT vs. PGD100: Test Statistic: 34520.0, P-value: 3.593${e}^{-14}$
\item Wilcoxon Test for SegRMT vs. CW: Test Statistic: 830.0, P-value: 5.978${e}^{-78}$
\item Wilcoxon Test for SegRMT vs. FGSM: Test Statistic: 16726.0, P-value: 3.270${e}^{-41}$
\end{itemize}
The low p-values (< 0.05) for all comparisons indicate that the differences in performance between the metamorphic testing tool and each of the gradient-based methods are statistically significant. This suggests that the tool's effectiveness in reducing the model's IoU scores is not due to chance.\newline
To further quantify the magnitude of the difference between the metamorphic testing tool and the gradient-based methods, we calculated Cohen's d effect size. We separated the methods into two groups: the metamorphic testing tool in one group and the gradient-based methods (PGD10, PGD40, PGD100, CW, and FGSM) in another group. The result is as follows:
\begin{itemize}
\item Cohen's d for Tool vs. Gradient-Based Attacks: -0.641
\end{itemize}
The negative value of Cohen's d indicates that the metamorphic testing tool group has lower IoU scores than the gradient-based methods group. The absolute value of 0.641 suggests a medium to large effect size, indicating that the difference between the two groups is substantial and practically significant.\newline
In summary, the statistical analysis using the Wilcoxon signed-rank test and Cohen's d effect size provides strong evidence that the metamorphic testing tool is more effective than the gradient-based adversarial attack methods in reducing the model's IoU scores. The differences are both statistically significant and practically meaningful, highlighting the tool's potential for robustness testing of deep learning models in semantic segmentation tasks.

\begin{table*}[h]
\centering
\caption{Performance as mIoU(\%) of Fine-tuned Models on Adversarial and Clean Datasets}
\label{tab:fine_tuned_performance}
\begin{tabular}{lcccccc}
\toprule
\textbf{Model} & \textbf{Clean} & \multicolumn{5}{c}{\textbf{Adversarial Testing Datasets}} \\ 
 & \textbf{Dataset} & \textbf{SegRMT} & \textbf{C\&W} & \textbf{FGSM} & \textbf{PGD10} & \textbf{PGD40} \\ \midrule
SegRMT   & \textbf{77.4}\%  & \textbf{53.8}\%  & 68.0\%  & 49.5\% & 45.0\%  & 46.0\%  \\ 
C\&W   & 76.9\%  & 9.8\%   & \textbf{72.0}\%  & 53.0\%   & 48.6\% & 51.0\%  \\ 
FGSM   & 76.8\%  & 10.0\%    & 72.0\%  & \textbf{66.0}\%   & 66.0\%  & 68.0\%  \\ 
PGD10  & 76.5\% & 4.0\%   & 72.0\%  & 65.0\%   & \textbf{68.0}\%  & 68.0\%  \\ 
PGD40  & 76.3\% & 2.0\%   & 73.0\%  & 62.0\%   & 66.0\%  & \textbf{66.0}\%  \\ \bottomrule
\end{tabular}
\end{table*}

\subsubsection*{RQ2: How effectively does SegRMT enhance seg-
mentation model robustness in self-adversarial testing
compared to traditional gradient-based attacks? \newline}

\paragraph{\textbf{Motivation:}}
This research question aims to evaluate which fine-tuned model achieves greater robustness: one fine-tuned using adversarial examples generated by SegRMT, and another fine-tuned using adversarial examples from traditional gradient-based techniques. Specifically, we want to determine if SegRMT’s adversarial examples lead to a more robust model when the same attack technique is used to challenge the model post-finetuning.

\paragraph{\textbf{Approach:}}
We conducted adversarial training on the \\ DeepLabV3 model using adversarial examples generated by various methods, including SegRMT, C\&W, FGSM, and PGD. The training dataset was augmented with these adversarial examples alongside the original clean images to challenge the model with a broad spectrum of perturbations. We implemented a systematic training protocol, as described in the Hyperparameter Selection section, which includes information on learning rates, batch sizes, and the proportion of clean to adversarial samples. The model underwent training for more than 80,000 iterations, during which its performance was assessed on both clean and adversarial validation sets. This approach allowed for a thorough evaluation of the model’s robustness against the specific adversarial attacks on which it was fine-tuned.

\paragraph{\textbf{Results:}}
The results presented in Table \ref{tab:fine_tuned_performance} highlight the significant improvements in robustness achieved through adversarial training. The model fine-tuned using C\&W adversarial examples exhibited the highest performance increase, with the mIoU improving from 21.7\% to 72\% on the C\&W-generated adversarial dataset. This substantial improvement demonstrates the effectiveness of fine-tuning the model using C\&W attacks, significantly enhancing its resilience to this particular form of attack. Similarly, the FGSM fine-tuned model showed a notable increase in performance, with the mIoU rising from 11.3\% to 66\%. The PGD10 and PGD40 fine-tuned models also demonstrated considerable improvements, with mIoU values increasing from 9.4\% to 68\% and from 8.5\% to 66\%, respectively.

The model fine-tuned using adversarial examples created by the SegRMT approach also showed a significant enhancement in performance, with the mIoU improving from 6.4\% to 53.8\%. While this increase is not as large as that achieved with the C\&W fine-tuning, it still demonstrates the overall effectiveness of SegRMT in improving the model’s resilience to its own perturbations. These results indicate that adversarial training, regardless of the method used to generate the adversarial examples, substantially improves the robustness of the model. However, the C\&W fine-tuned model showed the most substantial improvements, suggesting that this method may offer a particularly effective strategy for adversarial training when the goal is to enhance resistance to specific, well-optimized attacks.

\subsubsection*{RQ3: How effectively does SegRMT improve segmentation model robustness in cross-adversarial testing versus traditional gradient-based attacks?\newline}

\paragraph{\textbf{Motivation:}}
The aim of this research question is to evaluate the generalization capability of models fine-tuned with SegRMT-generated adversarial examples when tested across a variety of adversarial datasets. Cross-adversarial testing is essential for assessing whether a model’s robustness extends beyond the specific attacks it was trained on, thereby demonstrating its ability to defend against a broader range of adversarial perturbations.

\paragraph{\textbf{Approach:}}
Following the fine-tuning of the DeepLabV3 model using adversarial instances produced by SegRMT, together with conventional gradient-based approaches such as C\&W, FGSM, and PGD, we assessed the overall performance of these models on various adversarial datasets. The evaluation aimed to determine whether the robustness gained from training with one type of attack could generalize to other, dissimilar attacks. The specifics of the training process, including hyperparameters and the ratio of clean to adversarial examples, are detailed in the Hyperparameter Selection section. The model’s performance was evaluated by measuring its Mean Intersection over Union (mIoU) and Peak Signal-to-Noise Ratio (PSNR), specifically examining its ability to adapt robustly to both gradient-based and non-gradient-based adversarial examples.

\paragraph{\textbf{Results:}}
When analyzing the general performance of the fine-tuned models across various adversarial datasets, a clear pattern emerges. Models fine-tuned on adversarial examples generated by gradient-based methods, such as FGSM, PGD, and C\&W, generally performed well on other gradient-based adversarial datasets. This consistency is likely due to the shared characteristics and similar perturbation patterns among these attacks. However, these same models exhibited poor performance when tested against adversarial examples generated by SegRMT, a metamorphic approach. For instance, models fine-tuned with PGD10 and PGD40 saw their performance on SegRMT-generated adversarial examples degrade significantly, with mIoU dropping from 6.4\% in the base model to 4\% and 2\%, respectively.

Conversely, the model fine-tuned with SegRMT adversarial examples demonstrated better general performance across all types of attacks, including those generated by gradient-based methods. This suggests that the metamorphic testing approach produces more diverse and realistic adversarial examples, enhancing the model’s robustness against a broader range of perturbations. The results also indicate that while fine-tuning with specific gradient-based attacks improves resilience against similar perturbations, it may inadvertently reduce robustness against more diverse adversarial examples, such as those generated by SegRMT.

Additionally, the overall performance of the models on clean data experienced only a slight decrease following adversarial training. This minor degradation is a typical outcome in adversarial machine learning, where the trade-off for increased robustness against attacks is a small decrease in accuracy on non-adversarial inputs. The minimal decrease observed suggests that the adversarial training process effectively balanced the need for robustness with the maintenance of performance on clean data.

\begin{figure}[h]
\centering
\includegraphics[width=0.46\textwidth]{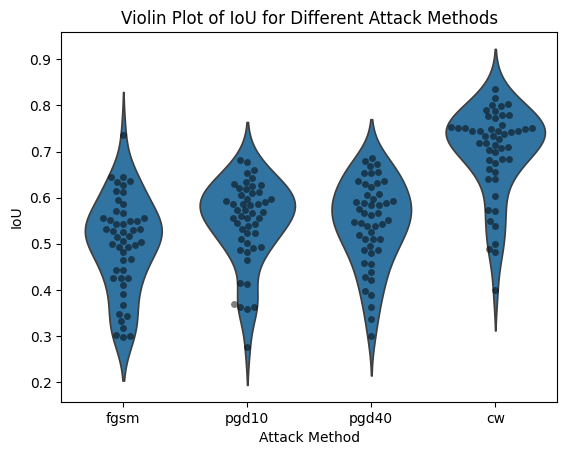}
\caption{Violin Plot of IoU for Different Attack Methods}
\label{fig:violin_plot}
\end{figure}

To determine the statistical significance of the performance disparities between the metamorphic testing tool and the gradient-based adversarial attack methods, we employed the Wilcoxon signed-rank test and calculated Cohen's d effect size.\newline
The Wilcoxon signed-rank test results consistently demonstrate that the tool's fine-tuned model significantly outperforms the gradient-based models on their respective adversarial datasets. The extremely low p-values (7.27${e}^{-12}$) obtained for all pairwise comparisons point to highly significant differences, implying that the tool's model exhibits greater robustness across a range of adversarial attack types.\newline
Cohen's d offers a quantitative gauge of the practical significance of the performance differences. The sizable negative Cohen's d value ($-$4.91) signifies a substantial effect size, wherein the tool's model demonstrates a marked improvement in performance over the gradient-based models. The magnitude of this effect size underscores the practical importance of the performance differences, above and beyond mere statistical significance.\newline
The confluence of Wilcoxon tests and Cohen's d provides compelling evidence that the tool's model is not only statistically superior to gradient-based models in terms of robustness against adversarial attacks but also practically superior, with a large effect size pointing to meaningful performance differences.\newline
The violin plot in Figure \ref{fig:violin_plot} lends further credence to this analysis by depicting the Intersection over Union (IoU) scores for various images derived from the adversarial examples generated by the gradient-based attack methods. The consistency in the performance of the model fine-tuned on adversarial examples generated by the metamorphic tool is readily apparent from the plot, as there are no significant outliers.\newline This finding indicates that the model maintains robust performance across a broad spectrum of adversarial examples, further emphasizing the effectiveness of the metamorphic tool in generating diverse and challenging adversarial examples that enhance the model's robustness.\newline
In sum, the metamorphic tool's capacity to generate varied adversarial examples renders it a valuable asset in preparing models for real-world adversarial scenarios, underscoring the importance of employing diverse adversarial training methods to achieve comprehensive robustness.

Overall, the metamorphic tool's ability to generate varied adversarial examples makes it a valuable asset in preparing models for real-world adversarial scenarios, underscoring the importance of employing diverse adversarial training methods to achieve comprehensive robustness.

\section{Discussion }
In RQ1, our findings reveal a significant distinction in the effectiveness of adversarial attacks generated by SegRMT compared to traditional gradient-based methods. Notably, SegRMT produced the lowest mIoU (6.4\%) while maintaining a higher PSNR of 24 dB, indicating that the adversarial examples generated by SegRMT are more detrimental to the model's performance. This suggests that SegRMT's adversarial examples, which include a broader range of realistic distortions—both perceptible and imperceptible—are particularly challenging for the model. The ability of SegRMT to introduce perturbations that retain a high degree of visual integrity while significantly degrading performance underscores its robustness as an adversarial testing approach.

In RQ2, further analysis of Table 2 results for the self-adversarial testing reveals that the C\&W attack achieves the highest mIoU (72\%) in self-adversarial testing, which correlates with its relatively modest impact on the model's performance, as seen in Table 1. The C\&W attack produces the least drop in mIoU , positioning it as the weakest attack among those evaluated. Weaker attacks like C\&W tend to generate higher mIoU scores in self-adversarial testing because they produce less impactful adversarial examples. This highlights a critical aspect of evaluating model robustness: weaker adversarial attacks may appear less detrimental in self-adversarial testing, but they also generate less challenging examples, which could lead to overestimated model robustness.

In RQ3, our findings also provide insights into the generalizability of the fine-tuned models. The results in Table 2 under the SegRMT column show that the model fine-tuned on SegRMT adversarial examples consistently outperforms others across all datasets, demonstrating superior robustness to various attacks, including those it has not encountered during training. This consistent performance underscores the robustness and versatility of SegRMT-generated adversarial examples, enabling the model to generalize better and withstand different types of perturbations. In contrast, models fine-tuned on gradient-based attacks, while performing well on adversarial datasets generated by other gradient-based methods, struggle significantly when tested on the SegRMT-generated dataset. This sharp decline in performance highlights a crucial weakness: gradient-based adversarial training enhances robustness against similar perturbations but fails to protect against more diverse, non-gradient-based adversarial examples.
In conclusion, SegRMT offers a more comprehensive and robust approach to adversarial testing and training, improving a model's ability to withstand a wide range of unseen adversarial examples and enhancing its overall robustness. On the other hand, while effective against similar attacks, gradient-based adversarial training does not offer the same level of protection against more sophisticated, non-gradient-based adversaries. Therefore, we suggest incorporating SegRMT into the adversarial training process to achieve a more generalized and robust model against a broader spectrum of adversarial threats.

\section{Threats to validity} \label{Threat}
As with any research study, it is important to carefully consider and address potential threats to the validity of the findings. Throughout the design and execution of our study, we have taken several steps to mitigate these threats and ensure the robustness and reliability of our results.
A critical consideration in comparative studies is ensuring fair evaluation between different methodologies. When comparing our metamorphic testing approach with baseline adversarial attack methods (FGSM, PGD, C\&W), we needed to control for factors that could unfairly advantage one method over another. To achieve this, we standardized the training environment by using identical parameter settings across all methods, including learning rate, batch size, and the ratio of clean to adversarial examples. This standardization serves two purposes: first, it ensures that performance differences arise from the inherent characteristics of each method rather than from advantageous parameter settings; second, it allows for meaningful comparison of computational efficiency and resource utilization across methods. Furthermore, we validated that these parameters were within the recommended ranges for each baseline method based on their original publications \cite{goodfellow2015,madry2017towards,carlini2017towards}
Another potential issue is the stochastic nature of the genetic algorithm (GA) used in our metamorphic testing approach. To mitigate the influence of randomness on our conclusions, we have employed a rigorous experimental design involving 10 runs with different random seeds and averaging results. This approach helps ensure that our findings are stable and reproducible, not influenced by chance.
Regarding the generalizability of our findings, we used a widely-accepted and representative dataset and model to conduct our study. The Cityscapes dataset is a standard benchmark for urban scene segmentation, and the DeeplabV3 model with a ResNet-50 backbone represents one of the top-performing models in the domain. While further testing on additional datasets and architectures would certainly be valuable, our choice of experimental materials provides a solid foundation for drawing meaningful conclusions about the effectiveness of our approach.
To facilitate reproducibility and enable other researchers to build upon our work, we have prioritized transparency in reporting our methodology and results. We have provided detailed information about our experimental setup, including hyperparameters and adversarial training procedures, and we have made our code and data publicly available. This allows for independent verification of our findings and promotes the accumulation of knowledge in the field.
Finally, while our approach has been specifically designed and evaluated in the context of robustness assessment for image segmentation models, we believe that it has significant potential for generalization to other domains. The modular structure of our framework and the flexibility of the metamorphic relations used suggest that our approach could be adapted to address similar challenges in tasks such as object detection or medical image analysis. This opens up exciting avenues for future research and highlights the broad impact and applicability of our work
\section{Conclusion and Future Work} \label{conc}
This paper presents a novel approach for evaluating and improving the robustness of image segmentation models. We achieve this through using metamorphic testing and adversarial training. The method we propose utilizes a genetic algorithm to create real-world perturbations and distortions. By including these adversarial examples in the training process, we have proven that our approach may significantly enhance the robustness and generalization of segmentation models against different forms of adversarial attacks.\newline
Our metamorphic testing approach has been demonstrated to outperform established adversarial attack approaches, such as FGSM, PGD, and C\&W, in decreasing model performance. This was achieved through extensive testing on the Cityscapes dataset utilizing the well-performed DeepLabV3 model. In addition, we have observed that models fine-tuned using our approach exhibit better generalization to unseen adversarial examples, highlighting the effectiveness of our method in improving overall model robustness. This emphasizes the effectiveness of our methodology in enhancing the general robustness of the model.
The results of this study have significant implications for the development and deployment of image segmentation models in real-world applications.
By providing a comprehensive and rigorous framework for assessing and enhancing model robustness, our approach helps ensure the reliability and the safety of these models in the face of various types of distortions and perturbations that may be encountered in the real-world.
This is particularly important in safety-critical domains such as autonomous driving and medical image analysis, where the consequences of model failures can be disastrous.
while our study has made significant contributions to the field of robustness assessment for image segmentation models, introducing a novel and unique approach for assessing and enhancing the robustness of image segmentation models using metamorphic testing and adversarial training There are several promising directions for future research that could further extend and improve upon our approach.
One such direction is the extension of our approach to other types of computer vision tasks and models.
While our current study focuses specifically on image segmentation using the Deeplabv3 model, the general principles of our approach could potentially be applied to other tasks such as object detection and image classification, other dataset types, such as HSI or remote sensing datasets and other model  architectures such as UNET or HRNet.
By demonstrating the generalizability of our method across different tasks and architectures, we could further establish its utility as a general-purpose framework for assessing and enhancing the models robustness.
Another potential avenue for future research is the integration of our approach with other techniques for improving model robustness, such as data augmentation, regularization, or architecture design. By combining our metamorphic testing and adversarial training approach with these complementary techniques, we may be able to achieve even greater improvements in model robustness and generalization.
In conclusion, our study introduces a novel and effective approach for assessing and enhancing the robustness of image segmentation models using metamorphic testing and adversarial training. The results of our experiments demonstrate the significant potential of this approach for improving the reliability and safety of these models in real-world applications. While there are many exciting opportunities for future research in this area, our study lays the foundation for further advances in the field of robustness assessment for image segmentation models. As we continue to refine and extend our approach, we remain committed to developing powerful and principled methods for ensuring the robustness and trustworthiness of these critical Deep learning systems.

\printcredits

\bibliographystyle{cas-model2-names}

\bibliography{cas-refs}





\end{document}